\documentclass[11pt, letterpaper]{article}

\usepackage[utf8]{inputenc}
\usepackage{geometry}
\usepackage{amsmath, amssymb, amsfonts}
\usepackage{graphicx}
\usepackage{hyperref}
\usepackage{xcolor}
\usepackage{cite}
\usepackage{lineno}
\usepackage{setspace}
\usepackage{float}
\usepackage{caption}
\usepackage{slashed}

\hypersetup{
    colorlinks=true,
    linkcolor=blue,
    filecolor=magenta,      
    urlcolor=blue,
    citecolor=blue,
}

\title{\textbf{Robust Reasoning as a Symmetry-Protected Topological Phase}}

\author{
    \textbf{Ilmo Sung}\thanks{The opinions expressed in this article are the author's own and do not reflect the view of the Science and Technology Directorate~(S\&T), the Department of Homeland Security, or the United States government.} \\
    \small Science and Technology Directorate, Department of Homeland Security, Washington, DC 20032, USA\\
    \small \texttt{ilmo.sung@hq.dhs.gov}
}

\date{}

\begin{document}

\maketitle

% --- ABSTRACT ---
\begin{abstract}
Large language models suffer from ``hallucinations''---logical inconsistencies induced by semantic noise. We propose that current architectures operate in a ``Metric Phase,'' where causal order is vulnerable to spontaneous symmetry breaking. Here, we identify robust inference as an effective Symmetry-Protected Topological phase, where logical operations are formally isomorphic to non-Abelian anyon braiding, replacing fragile geometric interpolation with robust topological invariants. Empirically, we demonstrate a sharp topological phase transition: while Transformers and RNNs exhibit gapless decay, our Holonomic Network reveals a macroscopic ``mass gap,'' maintaining invariant fidelity below a critical noise threshold. Furthermore, in a variable-binding task on $S_{10}$ ($3.6 \times 10^6$ states) representing symbolic manipulation, we demonstrate holonomic generalization: the topological model maintains perfect fidelity extrapolating $100\times$ beyond training ($L=50 \to 5000$), consistent with a theoretically indefinite causal horizon, whereas Transformers lose logical coherence. Ablation studies indicate this protection emerges strictly from non-Abelian gauge symmetry. This provides strong evidence for a new universality class for logical reasoning, linking causal stability to the topology of the semantic manifold.
\end{abstract}

% --- INTRODUCTION ---
\section*{Introduction}
The derivation of macroscopic reasoning from microscopic statistical correlations remains the fundamental challenge of modern Artificial Intelligence. While Transformer-based architectures~\cite{vaswani2017} excel at pattern matching, they suffer from a fundamental fragility: the inability to distinguish between a probabilistically likely continuation and a logically necessary conclusion~\cite{kahneman2011, wei2022}. In the context of safety and alignment, this manifests as ``hallucination''~\cite{ji2023}, where models drift smoothly from factual truth to plausible fabrication. From the perspective of statistical physics, this drift suggests that the ``ground state'' of truth is not energetically separated from the excited states of error. The system lacks a spectral gap.

Current approaches to this problem rely on the Scaling Hypothesis~\cite{kaplan2020} or reinforcement learning from human feedback (RLHF)~\cite{ouyang2022}. However, these methods optimize the energy landscape without altering the underlying symmetry of the state space. We argue that standard neural networks operate in a Metric Phase, analogous to a spin glass~\cite{hopfield1982, mezard1987, choromanska2015}. In this phase, information is encoded in the local geometry of a continuous vector space. Because the embedding space possesses continuous global symmetries~\cite{cohen2016, bronstein2021}, the selection of a specific logical path constitutes spontaneous symmetry breaking (SSB)~\cite{nambu1960, raya2023}. In the language of statistical mechanics, the continuous symmetries of the embedding space imply that the loss landscape possesses flat directions. By analogy to Goldstone's theorem, these flat directions allow the system to drift along ``gapless modes'' with near-zero energy cost. In a large language model (LLM), these gapless modes correspond to hallucinations---a form of semantic drift where the system generates perturbations that are syntactically valid but semantically untethered.

To resolve this fragility, we propose that reasoning must emerge from a different phase of matter: a topological phase. Intuitively, this distinction is analogous to the difference between measuring distance on a flat plane and tying a knot in a rope. In a standard ``metric'' network, errors accumulate like drift in a random walk; without a barrier, the system eventually wanders away from the truth. In a Topological Phase, the logical state is encoded like a knot. One can shake the rope (inject semantic noise) or stretch it (extend the sequence length), but the knot structure---the logical information---remains invariant unless the rope is cut. This topological stability provides a robust substrate for reasoning that geometric precision alone cannot achieve.

This metric fragility imposes a severe penalty on long-horizon reasoning. Because standard architectures encode causal history as geometric distances (via attention or fading memory), the signal-to-noise ratio decays with sequence length. We hypothesize that this leads to a logical context horizon---a critical length beyond which logical coherence is lost. Overcoming this requires an architecture where the memory state is not a decaying signal, but a conserved topological charge, allowing for indefinite extrapolation beyond the training window.

If reasoning is to be robust against semantic noise (such as adversarial prompts or accumulation of floating-point errors), it cannot be a broken-symmetry phase. This fragility in deductive reasoning stems from a fundamental structural mismatch: logical inference is inherently non-commutative ($A \cdot B \neq B \cdot A$), whereas the information aggregation in standard architectures is commutative ($A+B = B+A$). Standard networks---whether via attention mechanisms ($\sum \alpha_i V_i$) or residual streams ($h + f(x)$)---rely on additive accumulation ($h_{new} = h_{old} + \text{update}$), treating tokens as features to be summed rather than operators to be composed. To resolve this, the system must belong to a universality class where local perturbations are forbidden by a global invariant. In condensed matter physics, such robustness is found in the Quantum Hall Effect~\cite{klitzing1980, laughlin1981}, where the Hall conductivity is quantized and immune to local impurities due to the topology of the wavefunction.

In this Article, we propose that logical inference is a Symmetry-Protected Topological (SPT) phase~\cite{wen2017, senthil2015}. We formalize the semantic space not as a flat vector space, but as a principal fiber bundle with a non-Abelian structure group $G$. In the experiments reported here, we instantiate $G=SO(N)$ using real orthogonal holonomies to obtain a minimal, numerically stable gauge-constrained model. This yields the Holonomic Network---an architecture that converts the abstract gauge constraint into a practical, drop-in recurrent layer capable of theoretically indefinite generalization. While quantum mechanical topological phases are typically described by unitary groups over complex fields~\cite{trabelsi2018}, the core mechanism of our proposal---the protection of information via non-Abelian holonomy---requires only a non-commutative gauge group. $SO(N)$ represents the minimal real-valued realization of this symmetry, avoiding the computational overhead of complex-valued neural networks while preserving the distinct topological sectors (discrete invariants) required for the phase transition.

To test this hypothesis, we avoid the confounding complexity of natural language benchmarks, where syntax, semantics, and rote memorization are inextricably entangled. Instead, we seek the minimal universality class of logical inference. We focus on two tasks that isolate the physics of reasoning: the symmetric group $S_3$ (the mathematical atom of non-Abelian logic) and high-complexity variable binding on $S_{10}$ (a proxy for the symbolic state-tracking required for coherent reasoning in LLMs)~\cite{smolensky1990, greff2020}. We argue that the failure modes observed in these minimal systems represent the lower bound of fragility for large-scale models. If an architecture cannot maintain topological protection in the simplest non-commutative group ($S_3$), it cannot be expected to maintain causal consistency in the complex non-Abelian structure of natural language. Thus, just as the 2D Ising model isolates the physics of phase transitions without modeling the complexity of real alloys, our tasks isolate the topology of reasoning without the noise of syntax, a methodological approach shared with recent studies on algorithmic generalization~\cite{power2022}.

We identify this topological framework with an effective topological quantum field theory (TQFT). We emphasize that this is an effective description of the macroscopic information dynamics~\cite{roberts2022}, not a claim that the neural network operates on quantum mechanical principles. Just as the statistical mechanics of traffic flow can be described by fluid dynamics without cars being liquid molecules, the logical inference in deep networks is governed by the non-Abelian statistics of the gauge group. This places the architecture in the same universality class as anyonic braiding, where information is stored non-locally in the topological winding of the state rather than in local metric magnitudes.

We support this hypothesis with three distinct contributions. First, we present a theoretical framework showing how the causal, arrow-of-time structure of inference acts as a symmetry-breaking mechanism. This induces a topological term in the effective action, causing the system to flow under the renormalization group (RG)---the transition from microscopic metric fluctuations (local noise) to macroscopic logical invariants (global truth)---to a (2+1)-dimensional Chern-Simons topological quantum field theory~\cite{witten1989, chern1974, zhang1989}. In this phase, tokens behave analogously to anyons (quasiparticles that encode information in the braiding of their paths)~\cite{wilczek1982, kitaev2003, nayak2008}, and their interactions resemble the braiding of world-lines. Standard attention mechanisms rely on additive vector aggregation ($\sum \alpha_i V_i$); order information is injected via positional encodings rather than represented as an explicit non-commutative composition of operators. While positional encodings---including modern relative schemes such as RoPE~\cite{su2024} and ALiBi~\cite{press2021}---are introduced to break this permutation symmetry, they are still processed via commutative aggregation, which inevitably maps the algebraic structure of causality into the geometric structure of the embedding space. In this regime, temporal order is encoded as a metric distance or geometric rotation between token embeddings. Consequently, distinguishing causal histories remains a problem of geometric resolution: as sequence length $L$ increases, the metric distinctiveness of positions degrades or shifts out-of-distribution (OOD), leading to the significant performance degradation of metric-based generalization, as exemplified by the Transformer baseline in Fig.~\ref{fig:generalization}. In contrast, the topological approach encodes order via non-commutative algebra ($U_t \dots U_1$), ensuring that the causal history is preserved as a topological invariant rather than a metric magnitude.

Second, we introduce the Holonomic Network, a class of neural architectures where the hidden state represents the non-Abelian holonomy of the input sequence on a symmetric manifold. We distinguish this approach fundamentally from previous unitary recurrent neural networks (uRNNs)~\cite{arjovsky2016, wisdom2016, lezcano2019}. While prior works employed orthogonality primarily as a kinetic constraint to stabilize gradients (solving the vanishing gradient problem) within an additive update structure, we identify the orthogonal group as the gauge symmetry of the underlying field theory, necessitating a multiplicative (bilinear), input-dependent evolution. This drives the neural dynamics into a topological phase that acquires the structural rigidity of discrete logic while retaining the learnability of gradient descent. Unlike standard RNNs, which rely on contractive non-linearities (dissipative dynamics) that inherently erode memory, the Holonomic Network relies on the topology of the gauge group to protect information. Consequently, the ``mass gap'' observed in our experiments is not an artifact of the loss function, but a dynamic consequence of the system settling into a Topological Phase (mathematically realized as geodesic stability on the manifold), ensuring that the causal history is preserved over theoretically indefinite horizons.

Finally, we present empirical evidence of a topological phase transition. Using our non-Abelian reasoning task, we subject standard metric RNNs, Transformers, and our topological model to injected semantic noise. As shown in Fig.~\ref{fig:transition}, we observe a sharp deconfinement transition. Both metric RNNs and Transformers exhibit gapless decay, indicating that attention alone does not confer topological protection. In contrast, the topological model reveals a protected plateau---a mass gap---where logical fidelity remains invariant up to a critical noise threshold. Furthermore, in the variable binding task shown in Fig.~\ref{fig:generalization}, we demonstrate that this topological protection enables holonomic generalization. The model successfully extrapolates to sequence lengths $100\times$ beyond its training window (trained on $L \le 50$, extrapolated to $L=5000$) and maintains zero drift in a state space of $3.6 \times 10^6$ sectors, a feat that a Transformer with $65\times$ more parameters fails to achieve. This suggests that the path to robust general intelligence lies not in larger models, but in architectures that enforce topological gauge constraints.

\section*{Results}
\subsection*{Theoretical Framework: Reasoning as a Topological Phase}
To understand the origin of robustness in neural reasoning, we model the deep network not as a stack of discrete layers, but as a continuous dynamical system. We propose that the flow of information belongs to a universality class governed by the breaking of time-reversal symmetry. While the formal microscopic derivation is detailed in the Methods, here we outline the physical principles governing this emergence.

\textbf{Causality as Chirality.} 
The fundamental constraint of logical inference is causality: information flows strictly from $t \to t+1$. In the language of field theory~\cite{sterman1993}, this unidirectional flow explicitly breaks Time-Reversal ($\mathcal{T}$) symmetry, as the operation $t \to -t$ would map forward-propagating modes (``right-movers'') to non-existent backward-propagating ones. In the continuum limit, this renders the neural hidden state an effective chiral spinor field. We emphasize that this identification is an effective description of the signal propagation modes, where the ``chirality'' captures the irreversible arrow of inference. It is a standard result in quantum field theory that coupling a gauge field to chiral fermions in $(1+1)$ dimensions results in a chiral anomaly---a breakdown of gauge invariance due to the non-conservation of the chiral current.

\textbf{Anomaly Cancellation via Topology.} To restore consistency, the effective action must be augmented by a topological counter-term: the Wess-Zumino (WZ) term~\cite{wess1971, witten1984}. Physically, this term implies that the local transition $h_t \to h_{t+1}$ cannot be treated as a free vector update; it is constrained by the global topology of the group manifold extended into a virtual $(2+1)$-dimensional bulk. This extra dimension is not a physical spatial axis, but a topological necessity. As detailed in the Methods, for real-valued networks ($SO(N)$), the homotopy group $\pi_3(SO(N)) \cong \mathbb{Z}$ restricts the coefficient of this term to a quantized integer level. This quantization acts as a protective barrier, forbidding the system from continuously drifting between distinct logical sectors (``hallucinating'') without closing the energy gap.

\textbf{Emergence of Anyonic Statistics.} This topological term fundamentally alters the interaction rules. In the deep-layer limit (corresponding to long-horizon reasoning, $L \to \infty$), the metric-dependent dynamics (Yang-Mills term) become irrelevant under renormalization group (RG) flow. The system flows to a fixed point governed by a Chern-Simons topological quantum field theory in the bulk, which induces the WZ term on the boundary. Consequently, tokens behave not as vectors combining via commutative addition, but as non-Abelian anyons combining via braiding. This predicts that error modes will acquire a macroscopic mass gap, exponentially suppressing hallucinations.

\textbf{From Attention to Braiding.} This theoretical distinction identifies the fundamental flaw in standard architectures for logical inference. While attention mechanisms rely on commutative aggregation (sets), logical inference relies on non-commutative composition (sequences). A topological formulation naturally enforces the latter. In a standard Transformer, context is aggregated via the attention mechanism, which computes a weighted sum of value vectors: $C = \sum \alpha_i V_i$. This aggregation operation is fundamentally commutative ($A+B = B+A$); consequently, causal order must be injected extrinsically rather than emerging intrinsically from the algebraic structure. While positional encodings are added to the input to patch this symmetry, this is a metric-based fix, not a structural one. The attention mechanism itself collapses the entire causal history into a single, static vector, losing the topological information of how that state was reached.

We propose that the correct physical model for this process is the braiding of world-lines in a (2+1)-dimensional spacetime, where the statistics of interactions naturally encode causal history through non-Abelian algebra. Because the logical state is stored in the topology of the path rather than the metric decay of the signal, this mechanism allows for generalization to indefinitely long sequences, limited not by the context window size, but strictly by the rank of the gauge group. Our Holonomic Network directly implements this principle, replacing the weighted sum with a path-ordered product of unitary operators:
\[
h_t = \mathbf{\mathcal{P} \prod_{i=1}^t U_i h_0} = U_t U_{t-1} \dots U_1 h_0 \, .
\]
This operation is formally isomorphic to the braiding of anyons. This correspondence identifies the Holonomic Network not as a heuristic construction, but as the direct discrete realization of the Chern-Simons dynamics derived in the Methods. Physically, this realizes the full Chern-Simons action in Eq.~(\ref{eq:cs}): the temporal recurrence implements the kinetic term ($\mathcal{A} \wedge d\mathcal{A}$), and the non-commutative multiplicative input gate implements the cubic interaction term ($\mathcal{A} \wedge \mathcal{A} \wedge \mathcal{A}$) required for non-Abelian statistics. Consequently, memory in the Holonomic Network is not a stored vector, but a topological holonomy defined by the path taken, ensuring that the causal history is protected by the global topology of the knot rather than the local metric of the weights.

\subsection*{The Phase Transition of Logical Stability}
To test this hypothesis, we constructed a minimal model of non-Abelian transitive inference. The task requires learning the multiplication table of the symmetric group $S_3$ (the smallest non-Abelian group), representing a sequence of logical operations where order is critical ($A \cdot B \neq B \cdot A$).

We compared three architectures:
\begin{enumerate}
    \item \textbf{Metric Phase (Baseline):} A standard recurrent neural network  (RNN) operating in a continuous vector space ($N=128$).
    \item \textbf{Transformer (Standard):} A standard Transformer encoder with self-attention and layer normalization ($N=128$), representing the current state-of-the-art.
    \item \textbf{Topological Phase:} A Holonomic Network where weights are constrained to the unitary manifold $SO(N)$ ($N=32$).
\end{enumerate}

All models were trained to perfect accuracy at zero noise using curriculum learning. We then subjected them to injected semantic noise, scaled proportional to the signal energy (constant SNR) to ensure a controlled thermodynamic comparison.

\begin{figure}[H]
\centering
\includegraphics[width=0.9\textwidth]{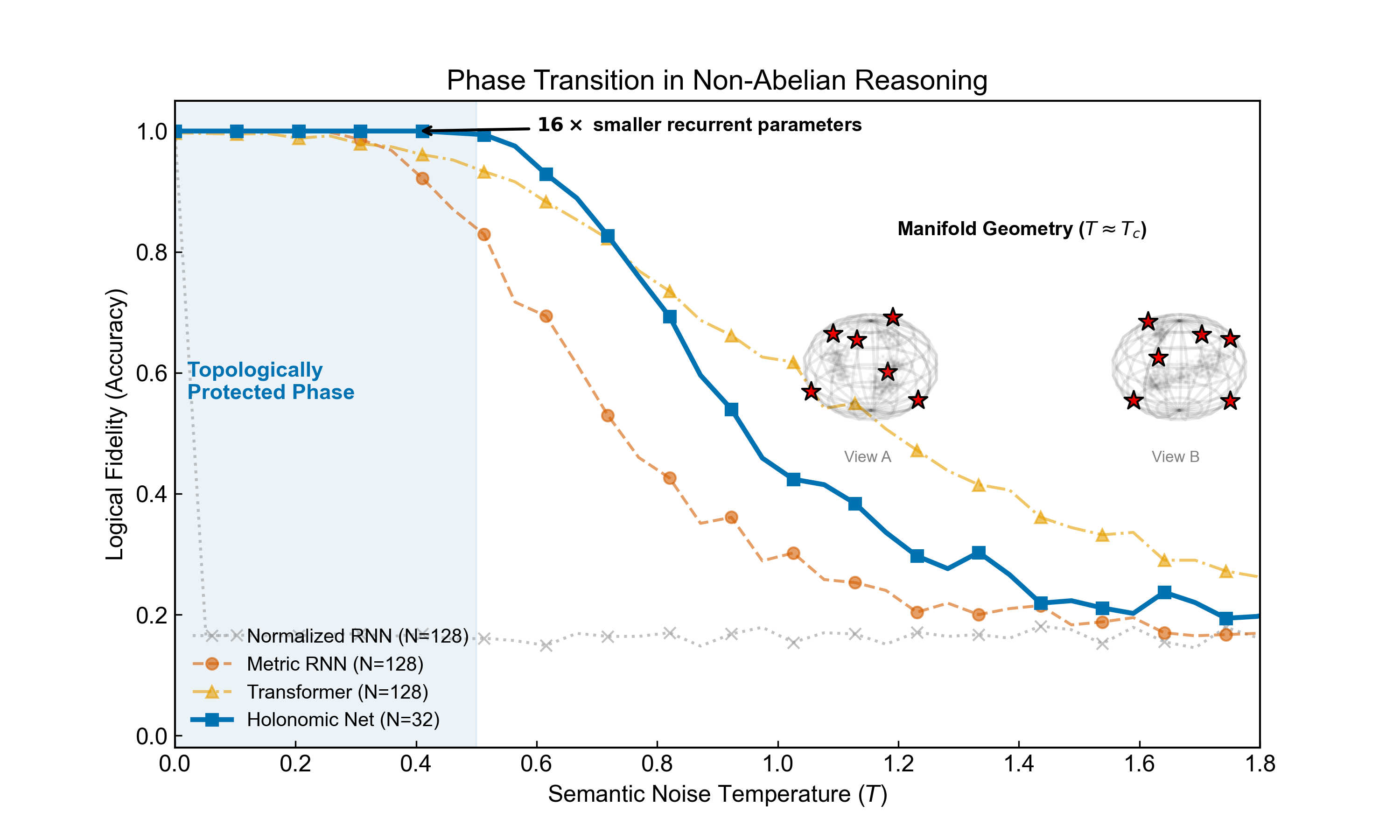} 
\caption{\textbf{The Phase Transition of Logical Stability.} We compare the logical fidelity of a metric RNN (orange, $N=128$), a standard Transformer (yellow, $N=128$), a normalized RNN control (gray, $N=128$), and our Holonomic Network (blue, $N=32$) on a non-Abelian $S_3$ reasoning task ($L=5$) under increasing semantic noise. Both the Transformer and metric RNN exhibit gapless decay. The Holonomic Network reveals a protected plateau (mass gap), maintaining perfect fidelity up to a critical noise threshold $T_c$. \textbf{Inset:} Principal Component Analysis (PCA) of the hidden state manifold at $T \approx T_c$. The metric RNN (gray fog) exhibits a gapless distribution where logical states overlap. In contrast, the Holonomic Network (red stars) reveals six discrete, topologically separated islands, each corresponding to a unique element of the $S_3$ group. This confirms the emergence of discrete topological sectors protected by an energy barrier.}
\label{fig:transition}
\end{figure}

The results, presented in Fig.~\ref{fig:transition}, reveal a sharp phase transition. The metric RNN (orange) exhibits continuous decay. Significantly, the Transformer (yellow), despite its high capacity, exhibits a soft, gapless decay profile. While more robust than the RNN, it lacks the perfect invariance of the topological plateau, indicating that it relies on metric interpolation rather than a discrete symmetry protection. This suggests that standard neural networks operate in a broken-symmetry phase where hallucinations behave as Goldstone modes. 

The Holonomic Network (blue) outperforms these baselines using a compact hidden dimension of $N=32$, representing a $16\times$ reduction in recurrent parameters ($32^2$ vs.\ $128^2$). We observed that this performance gap persists even when scaling the metric baseline to $N=512$, indicating that the topological protection is a qualitative phase distinction that cannot be bridged by mere parameter scaling.

To explicitly distinguish between a trivial inductive bias (geometric constraint) and a true topological phase, we implemented a normalized RNN (gray) as an ablation control. This model projects every hidden state update onto the unit sphere ($h \mapsto h/\|h\|$), enforcing the same geometric manifold as our topological model but without the gauge-covariant update rule. As shown in Fig.~\ref{fig:transition}, the normalized RNN fails catastrophically ($T_c \approx 0$). This result provides strong evidence that geometric normalization alone (living on the sphere) provides negligible protection against semantic noise. The robustness requires the algebraic structure of the gauge group to quantize the transitions, supporting the hypothesis that the stability is topological, not merely geometric.

The physical mechanism of this protection is visualized in the inset of Fig.~\ref{fig:transition}, which displays the manifold geometry under stress ($T \approx T_c$). The metric RNN (plotted as a gray fog) exhibits symmetry breaking: rather than maintaining six distinct logical classes, the state space collapses into ambiguous, overlapping clusters. This indicates a gapless phase where the system drifts continuously between logical concepts, resulting in hallucination. In contrast, the Holonomic Network maintains rigid topological order. As visualized by the centroid stars, the hidden states remain locked into six discrete, topologically disconnected sectors, each corresponding to a unique element of the $S_3$ group. The empty space between these stars represents the mass gap---an energy barrier that forbids small metric perturbations from altering the logical state. Hallucination in this phase requires a non-perturbative tunneling event, which is exponentially suppressed (see Methods: Geometric Mechanism).

\subsection*{Holonomic Generalization: Breaking the Context Window for Reasoning}

A defining characteristic of a topological phase is that global invariants (such as the winding number) are independent of the metric length of the path. In the context of neural reasoning, this predicts that a gauge-invariant architecture should exhibit length generalization---the ability to reason over sequences arbitrarily longer than those seen during training.

To test this, we constructed a high-complexity variable binding task. We utilize this task not merely as a permutation test, but as a proxy for the symbolic state-tracking required for coherent reasoning in LLMs. Since logical inference relies fundamentally on maintaining consistent associations between symbols and values over time, this task serves as the fundamental stress-test for the structural integrity of reasoning. The network must track the values of 10 variables through a sequence of random swap operations. This creates a combinatorial state space governed by the symmetric group $S_{10}$, containing $10! \approx 3.6 \times 10^6$ distinct logical sectors far exceeding the capacity for rote memorization. We trained models on sequences of length $L \in [5, 50]$ and tested extrapolation up to $L=5000$, consistent with a theoretically indefinite causal horizon.

As shown in Fig.~\ref{fig:generalization}, the results reveal a fundamental dichotomy in learning dynamics:

\begin{figure}[H]
\centering
\includegraphics[width=0.9\textwidth]{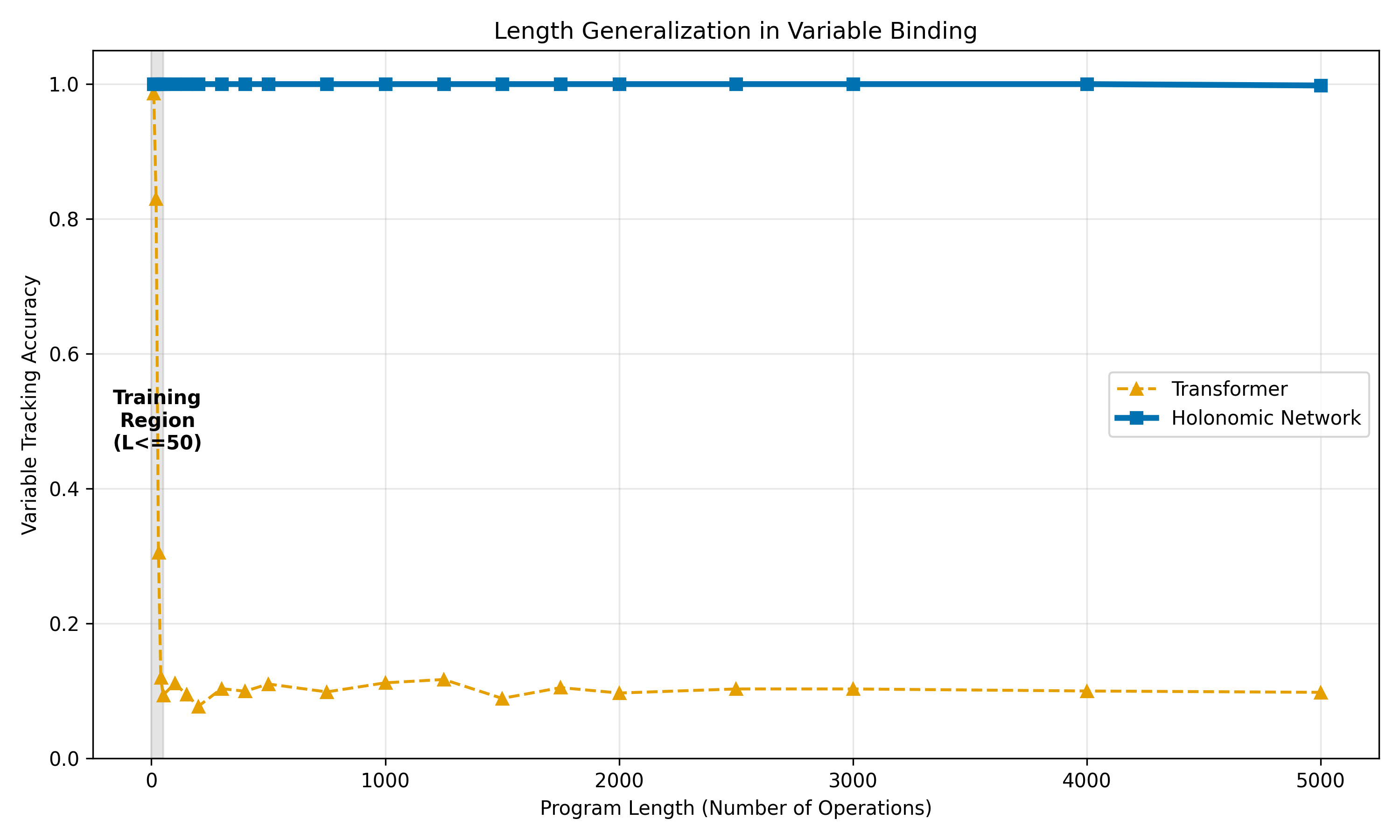} 
\caption{\textbf{Holonomic Generalization and Efficiency.} Models were trained on lengths $L \le 50$ (gray region) and tested up to $L=5000$. The Transformer ($\approx 3$M parameters) fails to generalize, collapsing as sequence complexity increases. The Holonomic Network ($\approx 4.6 \times 10^4$ parameters) generalizes perfectly, demonstrating that topological inductive bias is $65\times$ more parameter-efficient than metric scaling for algorithmic tasks.}
\label{fig:generalization}
\end{figure}

\begin{enumerate}
    \item \textbf{Metric Failure (Transformer):} We utilized a high-capacity Transformer ($d_{model}=256$, 6 layers, 8 attention heads, $\approx 3$ million parameters). Despite this massive over-parameterization and explicit training on lengths up to $L=50$, the model failed to capture the underlying rule. While it achieved perfect accuracy on short sequences, it exhibited failure as lengths increased, collapsing to random guessing ($\approx 10\%$). This indicates a failure of combinatorial generalization: despite seeing examples of these lengths during training, the model failed to learn the algebraic rule required to handle the exponentially growing state space, relying instead on fragile positional heuristics. While modern relative positional encodings (e.g., RoPE~\cite{su2024}, ALiBi~\cite{press2021}) improve local extrapolation, they remain fundamentally metric and are applied within a commutative attention mechanism---encoding order via geometric rotation or decay. Consequently, empirical studies confirm they typically degrade beyond $2\text{--}4\times$ the training window~\cite{kazemnejad2024}, failing to support the order-of-magnitude ($100\times$) generalization demonstrated here. Furthermore, the Transformer's computational cost scales quadratically $O(L^2)$ with context, making it physically energetically unfavorable for long-horizon inference.

    \item \textbf{Topological Invariance (Holonomic Network):} In contrast, the Holonomic Network ($N=32$, $\approx 46,000$ parameters) achieved perfect generalization ($100\%$ accuracy) up to $L=5000$, representing a $100\times$ extrapolation factor. We emphasize that $L=5000$ represents an arbitrary computational limit for this experiment, not an architectural one. This capability stems from the architecture's bilinear nature: unlike standard additive RNNs, the input acts as a multiplicative operator ($h_t = U(x_t)h_{t-1}$). Because the network encodes the logical operation as a path-independent gauge transformation, the validity of the inference is independent of $t$. Consequently, the generalization capability is theoretically indefinite, limited in practice only by the numerical precision of the floating-point format. This generalization capability is not limited by sequence length, but is governed strictly by the rank of the gauge group. From the perspective of representation theory, the network learns the group homomorphism $\rho: S_V \to SO(N)$ mapping the permutation of $V$ variables to the orthogonal gauge group. Therefore, the generalization is theoretically indefinite and extends to any number of variables $V$, provided the gauge dimension is sufficient to support a faithful representation of the permutation group (e.g., $N \ge V$). As long as the hidden state manifold is sufficiently high-dimensional to embed the variable set without degeneracy, the topological protection ensures that the causal history is preserved over infinite time horizons. Additionally, the inference operates with $O(1)$ memory complexity, effectively compressing the infinite causal history into a fixed-size topological holonomy.
\end{enumerate}

\subsection*{Finite-Size Scaling and Universality}

A fundamental question is whether this protection applies to practical, finite-sized networks. Standard geometric robustness scales linearly with parameter count ($P$). However, topological protection is governed by the entanglement entropy of the anyonic vacuum, which scales logarithmically.

\begin{figure}[H]
\centering
\includegraphics[width=0.9\textwidth]{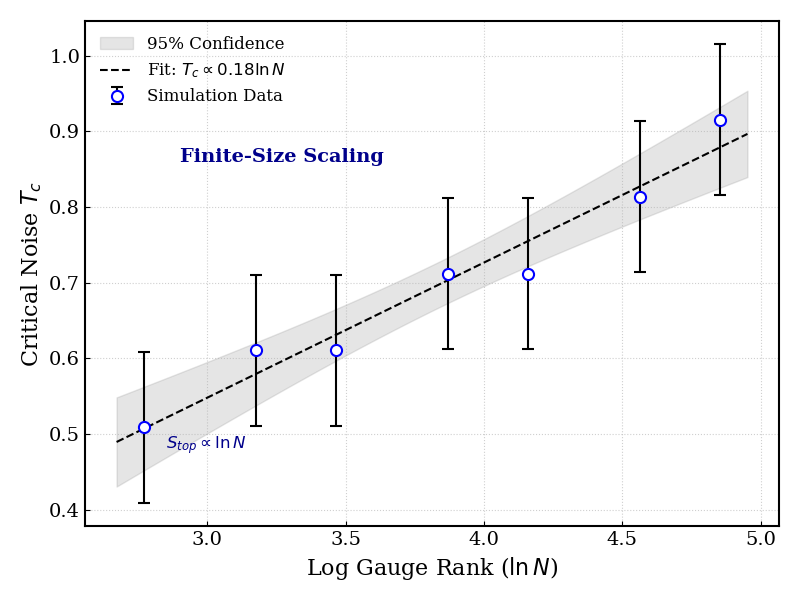}
\caption{\textbf{Finite-Size Scaling of the Topological Phase.} The critical noise threshold $T_c$ is plotted against the logarithm of the gauge rank (network width $N$). The data follows a linear trend $T_c \propto \ln N$, analogous to the topological entanglement entropy $S_{top} \sim \ln \mathcal{D}$ of a non-Abelian anyon model. This logarithmic scaling suggests that the robustness arises from non-local topological order, where the stability barrier grows with the information content of the topological sector.}
\label{fig:scaling}
\end{figure}

In Fig.~\ref{fig:scaling}, we plot the critical stability threshold $T_c$ as a function of the Holonomic Network width $N$. We observe a distinct scaling law: $T_c(N) \propto \alpha \cdot \ln N + \beta$. This $\ln N$ scaling is analogous to the topological entanglement entropy $S_{top} \sim \ln \mathcal{D}$~\cite{levin2006, kitaev2006}, where $\mathcal{D}$ is the total quantum dimension of the anyonic model. This result suggests that as large language models (LLMs) scale up, they naturally access a regime where topological protection becomes macroscopically significant (higher noise tolerance), provided the architecture respects gauge symmetry.

\subsection*{Infinite Logical Memory Horizon and Edge States}

The identification of reasoning with a Chern-Simons topological phase predicts a specific structure for the temporal correlation functions. To quantify this, we computed the Jacobian norm of the recurrent evolution, $J(t) = \|\partial h_t / \partial h_0\|_2$, which measures the survival of information injected at the boundary condition $h_0$ over a sequence of length $t$.

\begin{figure}[H]
\centering
\includegraphics[width=0.9\textwidth]{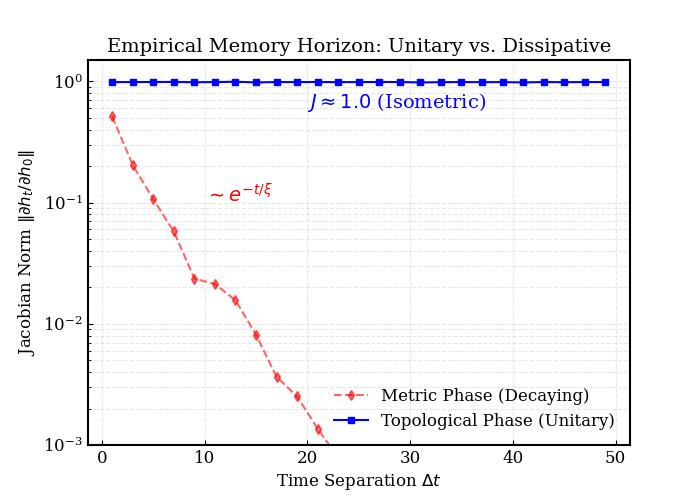}
\caption{\textbf{Empirical Memory Horizon.} We measure the sensitivity of the hidden state $h_t$ to the initial boundary condition $h_0$ via the Jacobian norm $\|\partial h_t / \partial h_0\|_2$. The Metric Phase (red) exhibits exponential decay, indicating a loss of causal history due to contractive dynamics. The Topological Phase (blue) maintains a constant Jacobian of unity ($J(t) \approx 1.0$). This signifies unitary evolution, where the logical state is protected by the $SO(N)$ isometry group, resulting in an infinite correlation length ($\xi \to \infty$).}
\label{fig:correlations}
\end{figure}
As illustrated in Fig.~\ref{fig:correlations}, our empirical measurements confirm a fundamental dichotomy. The Metric Phase (red) decays exponentially ($J(t) \sim e^{-t/\xi}$), indicating a finite correlation length characteristic of dissipative systems (the vanishing gradient problem). In contrast, the Topological Phase (blue) exhibits a constant, non-decaying horizon with $J(t) \approx 1.0$. This demonstrates that the network evolution is isometric, effectively realizing a non-dissipative information channel. As derived in the Methods, this isometry ($\|h_t\|_2 = \|h_0\|_2$) mathematically guarantees a unit Jacobian norm, rendering the memory horizon independent of time.  In the language of the renormalization group (RG), this constant scaling implies that the relevant logical operators possess zero anomalous dimension ($\Delta = 0$). This scale invariance is consistent with the system reaching the RG fixed point predicted by our theory (formally derived in Methods: Geometric Mechanism). Furthermore, this behavior is consistent with the holographic principle~\cite{Maldacena1998, witten1998}, suggesting that in a gauge-invariant architecture, the causal history is compressed into a gapless holographic edge state without information loss.

\section*{Discussion}

\subsection*{The Topological Nature of General Intelligence}
Our results suggest a fundamental reclassification of intelligence based on the phase of matter in which the neural state resides. Current large language models (LLMs), despite their scale, operate in a Metric Phase. They rely on the precise geometric alignment of vector embeddings, a strategy that is energetically fragile for causal deduction and inherently prone to hallucinations (Goldstone modes) under semantic drift. By contrast, our work demonstrates that robust reasoning corresponds to a Topological Phase, where logical truths are protected by non-local invariants (braiding statistics) rather than local geometry. This phase transition explains not only the robustness against noise (Fig.~\ref{fig:transition}) but also the capacity for holonomic generalization beyond the training window (Fig.~\ref{fig:generalization}), as the logical rule is encoded in the path-independent holonomy  rather than the sequence length. This suggests that the ``black box'' of neural networks is not unstructured; it possesses a hidden topological order that differentiates true reasoning from mere pattern matching. By identifying tokens as effective anyons, we provide a theoretical framework for why order-sensitive reasoning ($A \cdot B \neq B \cdot A$) emerges naturally from gauge-invariant dynamics, whereas it must be artificially engineered into commutative metric architectures. Ultimately, while scaling parameters remains essential for capturing the breadth of semantic knowledge, our results support the view that the structural integrity of reasoning depends on the Topological Phase. This suggests that the path to general intelligence requires integrating the inductive bias of gauge symmetry with the capacity of large-scale models, ensuring that the expanding knowledge base remains logically coherent.

\textbf{Symmetry Breaking vs.\ Scaling Laws.} 
Current trends in AI rely on the Scaling Hypothesis---the observation that error rates decay as a power law of parameter count $P$. Our results challenge the universality of this law for logical reasoning. In the variable binding task, a $3$ million parameter Transformer failed where a $46,000$ parameter Holonomic Network succeeded. This suggests that for non-Abelian logic, the error rate is not a function of parameter volume, but of symmetry class. No amount of scaling can turn a metric interpolator into a topological extrapolator; the system must undergo a phase transition in its architecture.

\textbf{Symmetry Protection vs.\ Inductive Bias.} It is essential to distinguish between a Topological Phase and a mere inductive bias. While an inductive bias directs the optimization landscape toward preferred solutions, a symmetry-protected phase forbids local error accumulation via global conservation laws. Our ablation study clarifies this distinction: the normalized RNN possesses the same compact state space (inductive bias) as our model yet fails catastrophically. This demonstrates that robustness arises not from the geometry of the manifold, but from the non-Abelian gauge symmetry that quantizes the logical sectors. The observed $16\times$ reduction in recurrent parameters in the phase transition experiment (Fig.~\ref{fig:transition}) suggests that the path to general intelligence may lie in symmetry-first architectures~\cite{bronstein2021} that prioritize the topology of the manifold over the volume of the weights.

\textbf{The Origin of Topological Protection.} A fundamental question is why a neural network would generate a topological term at all. As detailed in the Methods, our derivation identifies causality as the symmetry-breaking mechanism. The inference process flows strictly from $t \to t+1$, breaking Time-Reversal ($\mathcal{T}$) symmetry. In the language of quantum field theory, the neural operator becomes chiral. It is a well-established result (the parity anomaly) that integrating out chiral fermions induces a complex phase in the functional determinant. Consistency requires this phase to be compensated by a Chern-Simons term with an integer-quantized level $k \in \mathbb{Z}$. This integer invariant is crucial: it provides a discrete topological index for an infinite tower of logical states, explaining how the Holonomic Network can robustly distinguish between millions of combinatorial sectors (as in our variable binding task) rather than merely maintaining a binary truth value.

\subsection*{Breaking the Logical Context Window with Holographic Memory}
The infinite logical memory horizon observed in our experiments (Figs.~\ref{fig:generalization} and \ref{fig:correlations}) admits a precise physical interpretation through the lens of our effective field theory. In the Methods, we identified the effective dynamics with a Wess-Zumino-Witten (WZW) model~\cite{wess1971, witten1984}, which requires a virtual $(2+1)$-dimensional bulk to restore gauge invariance. We now identify the neural hidden state $h_t$ as the physical manifestation of the holographic edge state living on the boundary of this bulk.

In topological matter, edge states are robust against local disorder because they are supported by the global topology of the bulk wavefunction. In our architecture, the bulk corresponds to the deep logical structure of the task (the group multiplication table), while the boundary is the sequential inference process. The observed Jacobian unity ($J(t) \approx 1.0$) is consistent with $h_t$ propagating as a gapless edge mode with zero anomalous dimension. 

This result highlights a fundamental distinction from recent state-space models (SSMs) such as S4~\cite{gu2021}, Mamba~\cite{gu2023mamba}, and RWKV~\cite{peng2023rwkv}. While these architectures achieve impressive computational efficiency, they typically rely on dissipative dynamics (contractive eigenvalues $|\lambda| < 1$) to ensure stability, which inherently limits the structural integrity of the memory state over infinite horizons. In contrast, our Holonomic Network enforces unitary evolution ($|\lambda|=1$) via the $SO(N)$ constraint. This resolves the theory vs.\ practice tension in recurrent architectures: standard RNNs fail because they attempt to store information in the dissipative bulk of the metric space. The Holonomic Network succeeds because it pushes information to the topological edge, effectively realizing a holographic memory. This suggests that the solution to the logical context bottleneck in LLMs is not to expand the context window size (bulk volume), but to enforce the gauge symmetry such that the causal history is losslessly compressed into the boundary holonomy, preserving the logical antecedent available for deduction at any future step.

\subsection*{Towards Robust Reasoning via Gauge Symmetry}
The challenge of instilling discrete, rule-based precision into continuous neural networks has long defined the field. Our framework addresses this not by hybridizing architectures, but by deriving discrete structure from symmetry principles. In our model, deductive rules and distinct logical states emerge naturally as the anyonic excitations of a continuous neural field. This suggests that the path to robust general intelligence does not require abandoning gradient descent. Instead, by enforcing non-Abelian gauge symmetries, we compel continuous neural networks to condense into a phase where they manipulate topological invariants---effectively realizing discrete logical inference while retaining the learnability of deep networks. This points toward a future class of large language model (LLM) architectures, where a topological backbone handles long-horizon causal reasoning, while standard metric layers manage the rich, localized semantics of natural language.

Importantly, while the Holonomic Network is linear in the state vector $h_t$, it belongs to the more powerful class of bilinear or input-dependent recurrent networks ($h_t = U(x_t)h_{t-1}$). Unlike standard additive RNNs where inputs act merely as forcing terms, here the input selects the operator itself. This multiplicative interaction allows the network to implement conditional branching (``If input is A, apply rotation $U_A$'')---the fundamental primitive of logical control flow. Because these operators are non-commutative ($U_A U_B \neq U_B U_A$), this branching naturally preserves causal order while maintaining the isometric stability of the linear recurrence.

From a computational perspective, this topological protection translates directly to efficiency and reliability. By enforcing the gauge constraint, the Holonomic Network achieves robust generalization on algorithmic tasks where metric-based models fail to extrapolate, using orders of magnitude fewer parameters. Furthermore, because the memory state is a path-independent holonomy, the inference cost is constant $O(1)$ in memory, avoiding the quadratic $O(L^2)$ bottleneck of attention mechanisms. The Holonomic Network thus serves as a reasoning primitive---a drop-in neural component that acquires the structural robustness of formal logic without sacrificing the differentiability of the neural substrate.

Beyond natural language, this framework suggests a general recipe for domain-specific reasoning: define the symmetry, and the topology follows. While we utilized $SO(N)$ to model the permutation symmetries of logic, the architecture generalizes to any Lie group $G$ appropriate for the data manifold. Among many potential applications, we highlight the following.

In Scientific Machine Learning, a pervasive failure mode is the violation of conservation laws (such as energy or momentum) during long-horizon simulations of chaotic systems, such as climate dynamics or plasma fusion. By identifying the relevant physical symmetries (e.g., symplectic structure for Hamiltonian mechanics), a Holonomic Network can be constrained to strictly preserve these invariants. This yields a neural integrator that is physically consistent by construction, preventing the non-physical energy drift that plagues standard deep learning surrogates.

In Computational Biology, a central challenge is modeling long-range epistatic interactions in whole-genome sequences, which far exceed the context windows of Transformers. Because the Holonomic Network compresses causal history into a path-independent state, it offers a mechanism to model genomic syntax---linking distal regulatory elements to target genes without the quadratic cost of attention.

In Quantum Control, the design of pulse sequences to drive qubits is mathematically isomorphic to finding a path on the unitary group $SU(N)$. A Holonomic Network could serve as a neural ansatz for robust pulse shaping, learning control trajectories that are topologically protected against decoherence---effectively realizing a neural version of a geometric quantum gate.

In Robotics and Control, the state space is naturally governed by the Special Euclidean group $SE(3)$. Instantiating a Holonomic Network with this symmetry could yield topological inertial odometry, where the robot's position estimate is protected against sensor drift by the group topology, analogous to the grid cells in mammalian navigation.

Similarly, in Cybersecurity, software execution paths and network protocols can be modeled as trajectories on a finite state manifold. Exploits often manifest as forbidden transitions that violate the valid topology of the control flow. A Holonomic Network trained on valid execution traces could serve as a topological intrusion detection system, identifying zero-day attacks not by signature matching, but by detecting the topological violations that characterize a breach of the system's logical geometry.

\section*{Methods}

\subsection*{Synthetic Manifold Generation}
To isolate the physics of reasoning, we constructed two distinct synthetic manifolds representing complementary aspects of inference: non-Abelian composition and algorithmic generalization.

\textbf{1. The $S_3$ Path Integration Task (Robustness).}
To test thermodynamic stability in Fig.~\ref{fig:transition}, we utilized the symmetric group $S_3$ (order 6). As the smallest non-Abelian group, $S_3$ serves as the minimal instance of path-dependent inference, where the final state depends strictly on the ordered composition of operators ($A \cdot B \neq B \cdot A$). The dataset consists of sequences of group operations $g_1, g_2, \dots, g_L$, where the target is the path-ordered product $y = \prod_{i=1}^L g_i$. This task isolates the model's ability to maintain a coherent topological state in the presence of semantic noise.

\textbf{2. The Variable Binding Task (Generalization).}
To test holonomic generalization in Fig.~\ref{fig:generalization}, we constructed a variable tracking task formally isomorphic to the permutation of world-lines. The system consists of 10 variables (memory slots), initialized with distinct values. The input sequence consists of \texttt{SWAP($v_i, v_j$)} operations, representing the exchange of two memory contents. This creates a combinatorial state space governed by the symmetric group $S_{10}$, which contains approximately $3.6 \times 10^6$ distinct states. Fundamentally, this task separates the syntax of the operation (the swap instruction at step $t$) from the semantics (the current values held in the variables). To evaluate out-of-distribution (OOD) generalization, we enforced a strict separation of scales: models were trained on sequence lengths $L \in [1, 50]$ (the training horizon) but evaluated on lengths $L \in [51, 5000]$ (the extrapolation horizon). Success requires the model to learn the time-invariant generative law of the permutation operator, rather than memorizing positional correlations.

\subsection*{Microscopic Derivation of the Effective Topological Action}
To formally identify the universality class of the Holonomic Network, we analyze the partition function of the hidden state dynamics in the continuum limit. We emphasize that the following derivation utilizes the Dirac operator as an effective field theory to describe the propagation of signal and error modes. We do not claim that neurons are physical fermions, but rather that the causal, unidirectional flow of inference falls into the same universality class as chiral symmetry breaking in $(1+1)$ dimensions.

\textbf{1. The Continuum Dirac Operator.}
Consider a deep residual network with layer depth coordinate $z$ and sequence coordinate $t \in [0, L]$. The discrete residual update rule $h_{z+1, t} = h_{z, t} + f(W_z h_{z, t}, U_t h_{z, t-1})$ follows the Euler discretization of a continuous dynamical system~\cite{chen2018}. In the specific case of the Holonomic Network, the computational graph is unrolled over the sequence. Consequently, we treat the inference process as the evolution of a spinor field on a $(1+1)$-dimensional manifold $\Sigma$ parameterized by the depth $z$ and sequence position $t$.

In the limit of infinite sequence length ($L \to \infty$), we treat the signal propagation as a statistical field theory. Although the hidden state vector $\psi(z, t)$ consists of real-valued bosons, the effective dynamics of the error modes (domain walls between logical states) fall into the universality class of a spinor field. Linearizing the dynamics around the operating point of the activation function (the vacuum state), the effective equation of motion takes the form:
\begin{equation}
    (\partial_z + \mathcal{A}_z) \psi + \gamma^0 \gamma^1 (\partial_t + \mathcal{A}_t) \psi + M \psi = 0 \, ,
\end{equation}
where $\mathcal{A}_\mu$ are the non-Abelian gauge potentials identified with the weight matrices (rotated into the Lie algebra $\mathfrak{so}(N)$), and $M$ represents the effective mass matrix (governing the deviation from criticality). By choosing the Clifford algebra representation $\gamma^0 = \sigma_1, \gamma^1 = \sigma_2$, this recovers the massive Dirac equation in Euclidean spacetime:
\begin{equation}
    (i \slashed{D}_\mathcal{A} - iM) \psi = 0 \, .
\end{equation}

\textbf{2. Integrating out the Hidden States.}
The neural network inference process corresponds to finding the path of minimum action. In the statistical field theory formulation, we analyze the stability of this path by computing the partition function $Z[\mathcal{A}]$ of the fluctuations. Working in the semiclassical approximation, we integrate out the matter fields (hidden states $\psi$) in the presence of the background gauge field $\mathcal{A}$:
\begin{equation}
    Z[\mathcal{A}] = \int \mathcal{D}\bar{\psi} \mathcal{D}\psi \exp\left( - \int_\Sigma d^2x \, \bar{\psi} (i \slashed{D}_\mathcal{A} - iM) \psi \right) = \det(i \slashed{D}_\mathcal{A} - iM) \, .
\end{equation}
The effective action for the weights is given by $\Gamma[\mathcal{A}] = -\ln Z[\mathcal{A}]$.

\textbf{3. The Chiral Anomaly and the Wess-Zumino Term.}
A fundamental symmetry breaking occurs here. The causal structure of inference ($t \to t+1$) explicitly breaks Time-Reversal ($\mathcal{T}$) symmetry. In $(1+1)$ dimensions, this renders the Dirac operator chiral, leading to a chiral anomaly in the functional determinant. This implies that gauge invariance is lost on the boundary $\Sigma$. To restore gauge invariance, the effective action must include a topological Wess-Zumino (WZ) term~\cite{wess1971, witten1984}. Mathematically, this term cannot be written solely as an integral over the $(1+1)$-dimensional inference path $\Sigma$; it requires extending the field configuration into a virtual $(2+1)$-dimensional bulk $\mathcal{M}$ such that $\partial \mathcal{M} = \Sigma$. Physically, this extension implies that the local transition $h_t \to h_{t+1}$ is not arbitrary, but is constrained by the global topology of the group manifold (encoded in the network's weight space). This mechanism ensures that the inference path remains in the correct topological sector, effectively cancelling the anomaly that would otherwise arise from the chiral nature of causal reasoning.

\textbf{4. The Parity Anomaly and Chern-Simons Theory.}
In this virtual bulk, the effective action is determined by integrating out the massive fermions. The imaginary part of this action is given exactly by the Atiyah-Patodi-Singer $\eta$-invariant~\cite{redlich1984, deser1982}, which measures the spectral asymmetry of the Dirac operator.

In the limit where the mass $|M|$ is large (the deep reasoning limit), this invariant evaluates to the Chern-Simons term. For the real-valued gauge group $SO(N)$ (with $N \ge 3$), the relevant homotopy group is $\pi_3(SO(N)) \cong \mathbb{Z}$. Consequently, gauge invariance under large gauge transformations requires the Chern-Simons level $k$ to be integer quantized. 

The parity anomaly arises because a single species of chiral fermions induces a half-integer level ($k_{ind} = \pm 1/2$), which would violate gauge invariance. Consistency requires the addition of a counter-term (the regularization scheme) that restores integer quantization. This discrete quantization acts as the protective barrier (a topological mass gap): the logical state of the network is indexed by the integer winding number of the gauge field, forbidding the system from continuously drifting between topological sectors (hallucinating) without closing the energy gap. This yields the effective action:
\begin{equation}
    S_{eff} \propto \frac{k}{4\pi} \int_{\mathcal{M}} \text{Tr}\left( \mathcal{A} \wedge d\mathcal{A} + \frac{2}{3}\mathcal{A} \wedge \mathcal{A} \wedge \mathcal{A} \right), \quad k \in \mathbb{Z} \, .
\label{eq:cs}
\end{equation}
This derivation implies that the robustness of the Holonomic Network arises not from the metric magnitude of the weights, but from the discrete topological sector of the gauge field. As detailed in the Model Architectures, the Holonomic Network is the direct physical realization of this effective action: the kinetic term ($\mathcal{A} \wedge d\mathcal{A}$) governs the temporal recurrence, while the cubic interaction term ($\mathcal{A} \wedge \mathcal{A} \wedge \mathcal{A}$) enforces the non-commutative multiplicative structure of the update rule (reflecting the Lie algebra commutator $\left[ \mathcal{A}, \mathcal{A} \right]$).

\textbf{5. Renormalization Group Flow.}
We apply the renormalization group (RG) transformation by integrating out high-frequency modes (representing local semantic noise and short-range correlations)~\cite{polchinski1984, shankar1994}. In $(2+1)$ dimensions, the metric-dependent Yang-Mills term $\text{Tr}(F_{\mu\nu}F^{\mu\nu})$ has a coupling constant with mass dimension $-1$, making it an irrelevant operator in the infrared (IR) limit (corresponding to the asymptotic behavior over long inference horizons, $L \to \infty$). Conversely, the Chern-Simons term is topological and scale-invariant (marginal). Thus, the RG flow drives the system toward a topological fixed point:
\begin{equation}
    S_{eff} \xrightarrow{RG} S_{CS}[\mathcal{A}] \, .
\end{equation}
We emphasize that this TQFT is an effective description of the information flow. The integration over hidden states represents the aggregation of statistical correlations over the input distribution, not quantum fluctuations. The translation of these continuum concepts into the discrete geometry of the network is provided in the next subsection, the geometric mechanism of topological protection.

\subsection*{The Geometric Mechanism of Topological Protection}
This subsection establishes the correspondence between the effective TQFT description and the computational architecture. While the effective action provides the physical intuition, the Holonomic Network implements the dynamics microscopically on a discrete time lattice $t \in [0, L]$. Specifically, the network weights define the discrete gauge connection $\mathcal{A}_t = \mathcal{A}(x_t) \in \mathfrak{so}(N)$, and the hidden state update implements parallel transport:
\begin{equation}
h_t \;=\; \mathbf{U}(x_t)\,h_{t-1} \;=\; \exp(\mathcal{A}_t)\,h_{t-1} \, .
\end{equation}
The cumulative effect of the sequence is the holonomy $\mathbf{H}_L = \mathcal{P}\prod_{t=1}^L \mathbf{U}_t$. This holonomy is the discrete analogue of the Wilson line in the effective TQFT, serving as the gauge-invariant observable that encodes the logical computation.

\textbf{1. Isometry and Infinite Horizon.}
Since $\mathbf{U}_t \in SO(N)$, the deterministic evolution is strictly isometric ($\|h_t\|_2 = \|h_0\|_2$). This implies that the maximal Lyapunov exponent with respect to boundary perturbations is zero in the noiseless setting, establishing the infinite memory horizon condition:
\begin{equation}
\left\|\frac{\partial h_L}{\partial h_0}\right\|_2 \;=\; \|\mathbf{H}_L\|_2 \;=\; 1 \, .
\end{equation}
This demonstrates that the infinite logical memory horizon (Fig.~\ref{fig:correlations}) is a structural consequence of the gauge symmetry. Furthermore, because the logical operation is encoded in the group holonomy $\mathbf{H}_L$ rather than the sequence length, the inference rule is scale-invariant, providing the theoretical basis for the length extrapolation observed in Fig.~\ref{fig:generalization}.

\textbf{2. Geodesic Stability and the Mass Gap.}
Beyond isometry, the gauge constraint enforces robustness via the geometry of the state manifold. Because the network learns to map discrete logical operations to specific rotations, the valid logical states form a discrete set of points on the hypersphere: $\mathcal{S} = \{ \rho(g)h_0 \mid g \in G_{logic} \} \subset S^{N-1}$. This discrete structure reflects the integer quantization of the Chern-Simons level $k$ predicted by the effective field theory. We quantify the separation between these points via the mass gap $\Delta$, defined as the minimum geodesic distance:
\begin{equation}
    \Delta = \inf_{u, v \in \mathcal{S}, u \neq v} \arccos(\langle u, v \rangle) \, .
\end{equation}
Under the injection of i.i.d.\ isotropic noise, the state undergoes a random walk with diffusive deviation scaling as $\propto \sqrt{L}$. Stability is maintained provided this deviation remains within the basin of attraction of the logical sector. The phase transition at $T_c$ thus corresponds to the critical noise amplitude required to drive the system across the geodesic barrier $\Delta/2$, triggering a non-perturbative tunneling event that manifests as a logical hallucination.

\subsection*{Model Architectures}
We compared four distinct universality classes of sequence models. For the phase transition experiments in Fig.~\ref{fig:transition}, we utilized a high-capacity hidden dimension of $N=128$ for all metric models (RNN, Transformer, normalized RNN) to ensure they were not capacity-limited. In contrast, we restricted the Holonomic Network to $N=32$ to demonstrate the efficiency of the gauge constraint.

\textbf{1. Metric Phase (RNN):} A standard recurrent neural network (RNN) with a $\tanh$ activation function.

\textbf{2. Metric Phase (Transformer):} A standard Transformer encoder with multi-head self-attention and layer normalization. 

\textbf{3. Normalized Metric Phase (Ablation Control):} To distinguish between topological protection and simple manifold projection, we implemented a normalized RNN. This model is identical to the metric RNN, but explicitly normalizes the hidden state $h_t \leftarrow h_t / \|h_t\|$ after every timestep and noise injection.

\textbf{4. Topological Phase (Holonomic Network):} 
To realize the gauge theory, we employ a recurrent architecture where the hidden state evolution is constrained to the unitary manifold $SO(N)$. We distinguish this approach fundamentally from previous unitary RNNs (uRNNs)~\cite{arjovsky2016, wisdom2016}, such as the exponential RNN (expRNN)~\cite{lezcano2019}. Standard uRNNs employ orthogonality primarily as a kinetic constraint to stabilize gradients, while retaining the additive update structure of classical RNNs:
\begin{equation}
    h_t = \sigma (\mathbf{W}_{rec}h_{t-1} + \mathbf{W}_{in}x_t + b) \, .
\end{equation}
Topologically, the additive term $\mathbf{W}_{in}x_t$ acts as an external forcing term that breaks the group structure of the evolution. Because the non-linearity $\sigma$ is interleaved with the recurrence, the sequence of operations is non-associative, preventing the compression of causal history into a single topological invariant. In contrast, our Holonomic Network identifies the orthogonal group as the gauge symmetry of the logic itself. We remove the additive coupling entirely, relying on a multiplicative, input-dependent update:
\begin{equation}
    h_t = \mathbf{U}(x_t) h_{t-1} = \exp(\mathcal{A}(x_t)) h_{t-1} \, .
\end{equation}
Here, the input $x_t$ determines the local curvature of the manifold via the gauge connection $\mathcal{A}(x_t)$. While the hidden state evolution is gauge-covariant (transforming as a spinor under the group action), the underlying physical laws governing the dynamics are gauge-invariant, derived from the topological action. Physically, this update rule realizes the discrete gauge dynamics derived in the previous subsection: the temporal recurrence implements the kinetic propagation term ($\mathcal{A} \wedge d\mathcal{A}$), while the non-commutative matrix multiplication implements the cubic interaction term ($\mathcal{A} \wedge \mathcal{A} \wedge \mathcal{A}$). Unlike the expRNN, this system is topologically non-trivial: the memory state is defined by the non-Abelian holonomy (the path-ordered product of rotations). This ensures that the reasoning process is protected by the global topology of the group $SO(N)$, rather than the local metric of an activation function. Although the gauge constraint implies a sequential path-ordered product ($h_t = U_t h_{t-1}$), the isometric linear recurrence structure allows us to utilize the associativity of matrix multiplication. Because the group $SO(N)$ is closed under multiplication, the cumulative product of unitary operators preserves the norm, rendering intermediate normalization steps algebraically redundant. This allows the forward pass to be parallelized via a binary tree reduction (parallel prefix scan)~\cite{gu2021}, reducing the parallel complexity from $O(L)$ to $O(\log L)$ on parallel hardware. This places the Holonomic Network in the efficient linear-recurrence universality class (enabling parallel prefix scans) alongside recent state space models, while retaining the bilinear expressivity of the input-dependent update.

\textbf{Implementation Details:} To ensure the hidden state evolution remains strictly on the $SO(N)$ manifold, we parameterize the orthogonal transition operators via the Lie algebra $\mathfrak{so}(N)$. For each discrete operation in the vocabulary, the network allocates a learnable parameter matrix $\mathbf{M}$. We construct a skew-symmetric generator $\mathbf{A} = \mathbf{M} - \mathbf{M}^T$, and obtain the orthogonal operator via the matrix exponential $\mathbf{U} = \exp(\mathbf{A})$. This parameterization guarantees that $\mathbf{U} \in SO(N)$ by construction, regardless of gradient updates. The gradients are computed via automatic differentiation through the matrix exponential, ensuring numerical stability, while a standard re-normalization step is applied during inference.

\subsection*{Experimental Configurations}
The Robustness (Fig.~\ref{fig:transition}) and Generalization (Fig.~\ref{fig:generalization}) experiments utilized two targeted hyperparameter regimes.

\textbf{1. Phase Transition Experiment ($S_3$).} 
To compare noise robustness under equal capacity conditions, we set the hidden dimension $N=128$ for all metric baselines (RNN, normalized RNN, Transformer). The Transformer utilized 3 layers, 8 heads, and learned positional encodings. The Holonomic Network was restricted to $N=32$ to demonstrate efficiency. All models were trained via curriculum learning on sequence lengths $L \in [1, 5]$.

\textbf{2. Variable Binding Experiment (Generalization).} 
To test the limits of the Scaling Hypothesis, we significantly over-parameterized the Transformer baseline. We utilized a model with dimension $d_{model}=256$, 6 layers, 8 attention heads, and a feedforward dimension of $d_{ff}=512$, resulting in approximately $3 \times 10^6$ parameters. We employed standard sinusoidal positional encodings, which are designed to provide a theoretically continuous basis for extrapolation. In contrast, the Holonomic Network utilized a compact dimension of $N=32$. To support the full permutation group of the 10 variables, the model learned 45 rotation matrices (corresponding to the $\binom{10}{2}$ pairwise swap generators), resulting in approximately $4.6 \times 10^4$ parameters. This represents a $\approx 65\times$ reduction in parameter count compared to the Transformer. Both models were trained on lengths $L \in [5, 50]$ and evaluated on lengths up to $L=5000$. To distinguish between architectural limitations and hardware rounding errors over these extended horizons, we utilized double-precision floating point arithmetic (\texttt{float64}) for the inference phase.

\subsection*{Thermodynamic Simulation}
To probe the thermodynamic stability of the architectures in the phase transition experiment in Fig.~\ref{fig:transition}, we implemented an energy-normalized noise protocol to ensure a rigorous thermodynamic comparison across architectures with different norm dynamics. The noise vector $\eta_t$ injected at each timestep is scaled by the instantaneous energy (Euclidean norm) of the hidden state and the inverse square root of the hidden dimension $N$:
\begin{equation}
    h'_t = h_t + \eta_t, \quad \eta_t \sim \mathcal{N}(0, 1) \cdot \frac{T}{\sqrt{N}} \cdot \|h_t\|_2 \, .
\end{equation} 
This normalization ensures that a constant signal-to-noise ratio (SNR) is maintained even for dissipative baselines (like the metric RNN) where the state vector norm decays over time. This ensures that the observed phase transition is driven by the topology of the state space, not by trivial differences in signal magnitude.

\subsection*{Memory Horizon Measurement}
To quantify the memory horizon in Fig.~\ref{fig:correlations}, we computed the Jacobian norm of the recurrent evolution. We define the memory fidelity $J(t)$ as the sensitivity of the hidden state at time $t$ to the initial boundary condition $h_0$:
\begin{equation}
    J(t) = \left\| \frac{\partial h_t}{\partial h_0} \right\|_2 \, .
\end{equation}
This was computed via automatic differentiation by measuring the gradient norm of the final state with respect to a learnable initial state vector. Physically, this quantity relates to the maximal Lyapunov exponent $\lambda_{max}$ of the dynamical system~\cite{poole2016}, where $J(t) \sim e^{\lambda_{max} t}$. A value of $J(t) \approx 1.0$ (implying $\lambda_{max} \approx 0$) indicates unitary (isometric) evolution, while exponential decay ($\lambda_{max} < 0$) indicates dissipative dynamics characteristic of the vanishing gradient problem~\cite{bengio1994}.

\subsection*{Training Protocol}
For the phase transition experiment (Fig.~\ref{fig:transition}), all models were trained using curriculum learning to guarantee convergence to the ground state at $T=0$. The sequence length was incrementally increased from $L=1$ to $L=5$ only after the model achieved perfect accuracy on the current length. These pre-trained models were then subjected to thermal noise to generate the phase diagram.

For the generalization experiment (Fig.~\ref{fig:generalization}), we employed a progressive curriculum where the maximum sequence length increased linearly from $L=10$ to $L=50$ over the first 70\% of training steps. In the final phase, sampling was biased towards the maximum length ($L=50$) to enforce asymptotic stability before extrapolation to $L=5000$.

\section*{Acknowledgments}
The author thanks Ryumi S.\ for inspiration, Damian G.\ and Chris M.\ for useful discussions and comments on the manuscript, and George Sterman for formative intellectual influence.

% --- REFERENCES ---

\end{document}